%% file: main_paper_arxiv.tex
\documentclass[twoside,leqno,twocolumn]{article}

\usepackage[letterpaper]{geometry}

\usepackage{ltexpprt}
\usepackage{hyperref}

\usepackage{amsmath,amssymb,amsfonts}
\usepackage{adjustbox}
\usepackage{booktabs}
\usepackage{multirow}
\usepackage{subcaption}
\usepackage{xcolor}
\usepackage{algorithm}
\usepackage{algorithmic}
\usepackage{tabularx}
\usepackage{subcaption}
\newcommand\sbullet[1][.5]{\mathbin{\vcenter{\hbox{\scalebox{#1}{$\bullet$}}}}}
\newcommand{\name}{CoLafier}
\newcommand{\fullname}{CoLafier: \underline{Co}llaborative Noisy \underline{La}bel puri\underline{fier} with LID guidance}
\newcommand{\disname}{LID-dis}
\newcommand{\disfullname}{LID-based noisy label discriminator}
\newcommand{\puriname}{LID-gen}
\newcommand{\purifullname}{LID-guided label generator}

\allowdisplaybreaks
\begin{document}
\newcommand\relatedversion{}

\title{CoLafier: \underline{Co}llaborative Noisy \underline{La}bel Puri\underline{fier} With Local Intrinsic Dimensionality Guidance}
\author{Dongyu Zhang\thanks{Worcester Polytechnic Institute, 
\{dzhang5, rhu, rundenst\}@wpi.edu}
\and Ruofan Hu\footnotemark[1]
\and Elke Rundensteiner\footnotemark[1]}
\date{}

\maketitle


\fancyfoot[R]{\scriptsize{Copyright \textcopyright\ 2024 by SIAM\\
Unauthorized reproduction of this article is prohibited}}





\begin{abstract} \small\baselineskip=9pt 

\input{sections/sec0_abstract}
\end{abstract}


\input{sections/sec1_introduction}

\input{sections/sec2_related_work}

\input{sections/sec3_methodology}

\input{sections/sec4_experiment}

\input{sections/sec5_conclusion}

\input{sections/sec6_acknowledgement}

\bibliographystyle{siam}
\bibliography{references}

\input{sections/supp}
\end{document}

%% file: sections/sec0_abstract.tex
Deep neural networks (DNNs) have advanced many machine learning tasks, but their performance is often harmed by noisy labels in real-world data. Addressing this, we introduce \name{}, a novel approach that uses Local Intrinsic Dimensionality (LID) for learning with noisy labels. \name{} consists of two subnets: \disname{} and \puriname{}. \disname{} is a specialized classifier. Trained with our uniquely crafted scheme, \disname{} consumes both a sample's features and its label to
predict the label -
which allows
it to 
produce an enhanced internal representation. 
We observe that LID scores computed from this representation that effectively distinguish between correct and incorrect labels across various noise scenarios. In contrast to \disname{}, \puriname{}, functioning as a regular classifier, operates solely on the sample's features. During training, \name{} utilizes two augmented views per instance to feed both subnets. \name{} considers the LID scores from the two views as produced by \disname{}  to assign weights in an adapted loss function for both subnets. 
Concurrently, \puriname{}, serving as classifier,
suggests pseudo-labels.
\disname{} then processes these pseudo-labels along with two views to derive LID scores. Finally, these LID scores along with the differences in predictions from the two subnets guide the label update decisions. This dual-view and dual-subnet approach 
enhances the overall reliability of the framework. Upon completion of the training, 
we deploy the \puriname{} subnet of  \name{}  as the final classification model. \name{} demonstrates improved prediction accuracy, surpassing existing methods, particularly under severe label noise. For more details, see the code at \url{https://github.com/zdy93/CoLafier}.

%% file: sections/sec1_introduction.tex
\section{Introduction}

{\bf Motivation.} 
Deep neural networks (DNNs) have achieved remarkable success in a wide range of machine learning tasks \cite{zhang2020time,zhang2021human, zhang2023fata}. Their training typically requires extensive, accurately labeled data. However, acquiring such labels is both costly and labor-intensive \cite{song2022learning, zhang2021lancet, hofmann2022demonstration}. To circumvent these challenges, researchers and practitioners increasingly turn to non-expert labeling sources, such as crowd-sourcing \cite{hu-etal-2022-tweet} or automated annotation by pre-trained models \cite{song2019selfie}. Although these methods enhance efficiency and reduce costs, they frequently compromise label accuracy \cite{hu-etal-2022-tweet}. The resultant 'noisy labels' may inaccurately reflect the true data labels. Studies show that despite their robustness in AI applications, DNNs are susceptible to the detrimental effects of such label noise, which risks  impeding their performance and also generalization ability \cite{arpit2017closer,song2022learning}.

{\bf State-of-the-Art.}
Recent studies on learning with noisy labels (LNL) reveal that Deep Neural Networks (DNNs) exhibit interesting memorization behavior  \cite{arpit2017closer, xia2021robust}. Namely, DNNs tend to first learn simple and general patterns, and only gradually begin to learn  more complex patterns, such as data with noisy labels. Many methods thus leverage signals
from the  early training stage  
\cite{li2023disc}, such as 
loss or confidence scores,
to identify potentially incorrect labels.
For label correction, 
the identified faulty labels are either dropped,
assigned with a reduced  importance score, or replaced with generated pseudo labels 
\cite{han2018co,li2019dividemix, zhang2021elite}.

However, these methods can suffer from accumulated errors caused by incorrect selection or miscorrection - with the later further negatively affecting the representation learning and leading to potential overfitting to noisy patterns \cite{song2022learning, tu2023learning}. 
Worse yet,
most methods require prior knowledge about the noise label ratio or the specific pattern of the noisy labels \cite{han2018co, li2023disc}. In real-world scenarios, this information is typically elusive, making it difficult to implement these methods.

Local Intrinsic Dimensionality (LID), a measure of the intrinsic dimensionality of data subspaces \cite{houle2017local}, 
can be leveraged 
for  training DNNs  on noisy labels.
Initially, LID decreases as the DNN models the low-dimensional structures inherent in the data. Subsequently, LID increases, indicating the model's shift towards overfitting the 
noisy label. 
Another study \cite{Ma_Li_Wang_Erfani_Wijewickrema_Schoenebeck_Song_Houle_Bailey_2017} applied LID to identify adversarial examples 
in DNNs, which typically increase the local subspace's dimensionality. These findings suggest LID's sensitivity to noise either from input features or labels. Nonetheless, previous research has  utilized LID as  a general indicator for the training stages or for detecting feature noise.
While a promising direction for research,
applying  LID for detecting mislabeled samples has not been explored before.

{\bf Problem Definition.}
In this study,
we 
propose a method for solving  classification with noisy labeled training data. 
Given a set of training set with each item labeled with  one noisy classification label,
our goal is to train a robust classification model that solves the classification task accurately without any knowledge about the quality or correctness of the given labels.



{\bf Challenges.} Classification with noisy labeled training data is challenging for the following reasons:

$\sbullet[.75]$
\textit{Lack of knowledge about noise ratio and noise pattern.}
Without knowledge about the noise ratio and noise pattern of the given dataset, it is challenging to develop a universal method that can collect sufficient clean labels to train a strong model.

$\sbullet[.75]$
\textit{Compounding errors in the training procedure.} Incorrect selection or correction errors made early in the learning process can compound, leading to even larger errors as the model continues to be trained. This can result in a model that is  far off from the desired outcome.

{\bf Proposed Method.}
In response to these challenges, 
we conduct an empirical study to evaluate the effectiveness of the Local Intrinsic Dimensionality (LID) score as a potential indicator for mislabeled samples. We design a specialized classifier,  namely, 
\disfullname{} (\disname{}). \disname{} processes both a sample's features and label to predict the label. Notably, its intermediate layer yields an enhanced representation encompassing both feature and label information. Our uniquely crafted training scheme for \disname{} reveals that the LID score of this representation can effectively differentiate between correctly and incorrectly labeled samples. This differentiation is consistent across various noise conditions.

To complement \disname{}, we introduce the \purifullname{} (\puriname{}), a regular classification model that operates solely on the data's features - not requiring access to the label. \disname{} and \puriname{} together as two subnets form our proposed framework, \fullname{}. During training, we generate two augmented views of each instance's features, which are then processed by both \disname{} and \puriname{}. \name{} consider the consistency and discrepancy of the two views' LID scores as produced by \disname{} to determine weights for each instance in our adapted loss function. This reduces the risk of incorrect weight assignment. Both \disname{} and \puriname{} undergo training using their respective weighted loss. Concurrently, \puriname{} suggests pseudo-labels from these two augmented views for each training instance. \disname{}  processes these pseudo-labels along with two views, deriving LID scores for them. These LID scores and the difference between prediction from \disname{} and \puriname{} guide the decision on the label update. Information from the two views and two subnets together helps mitigate the risk of label miscorrection. After training is complete, \puriname{} is utilized as the  classification model to be deployed.

\textbf{Contributions.} Our contributions are as follows:


$\sbullet[.75]$
We craft a pioneering approach to harness the LID score in the context of noisy label learning, leading to the development of  \disname{} subnet. \disname{} processes not
only  a sample's features but also its label as input. This yields an enhanced representation adept at distinguishing between correct and incorrect labels across varied noise ratios and patterns.



$\sbullet[.75]$
Drawing insights from the LID score, we introduce the \name{} framework, a novel solution that integrates two \disname{} and \puriname{} subnets.  This framework utilizes two augmented views per instance, applying LID scores from the two views to weight the loss function for both subnets. LID scores from two views and the discrepancies in prediction from the two subnets inform the label correction decisions. This dual-view and dual-subnet approach significantly reduces the risk of errors and enhances the overall  effectiveness of the framework.

$\sbullet[.75]$
We conduct  evaluation studies across varied noise conditions.  Our findings demonstrate that, even in the absence of explicit knowledge about noise characteristics, \name{} still consistently yields improved performance compared to state-of-the-art LNL methods.


%% file: sections/sec2_related_work.tex
\section{Related Works}
\subsection{Learning With Noisy Labels.}
In recent studies, two primary techniques have emerged for training DNNs with noisy labels: sample selection and label correction. Sample selection approaches focus on identifying potentially mislabeled samples and diminishing their influence during training. Such samples might be discarded \cite{han2018co, wei2020combating}, given reduced weights in the loss function \cite{Ren_Zeng_Yang_Urtasun_2018, hu2023uce}, or treated as unlabeled, with semi-supervised learning techniques applied \cite{li2023disc, li2019dividemix}. On the other hand, label correction strategies aim to enhance the training set by identifying and rectifying mislabeled instances. Both soft and hard correction methods have been proposed \cite{Ren_Zeng_Yang_Urtasun_2018, song2019selfie, wu2021learning}. However, a prevalent challenge with these approaches is the amplification of errors during training. If the model makes incorrect selection or correction decisions, it can become biased and increasingly adapt to the noise. Another challenge arises when certain methods presuppose knowledge of the noise label ratio and pattern, using this information to inform their hyper-parameter settings \cite{han2018co,li2019dividemix,li2023disc}. However, in real-world scenarios, this information is typically unavailable, rendering these methods less practical for implementation.


\subsection{Supervised Learning and Local Intrinsic Dimensionality.}
The Local Intrinsic Dimensionality (LID) \cite{houle2017local} has been employed to detect adversarial examples in DNNs, as showcased by \cite{Ma_Li_Wang_Erfani_Wijewickrema_Schoenebeck_Song_Houle_Bailey_2017}. Their research highlights that adversarial perturbations, a specific type of input feature noise, tend to elevate the dimensionality of the local subspace around a test sample. As a result, features rooted in LID can be instrumental in identifying such perturbations. Within the Learning with Noisy Labels (LNL) domain, LID has been employed as a global indicator to assess a DNN's learning behavior and to develop adaptive learning strategies to address noisy labels \cite{ma2018dimensionality}. However, it has not been utilized to identify samples with label noise.

In contrast to these applications, our study introduces a framework that leverages LID to detect and purify noisy labels at the sample level. Using LID, we can differentiate between samples with accurate and inaccurate labels, and its insights further guide the decision to replace noisy labels with more reliable ones.

%% file: sections/sec3_methodology.tex
\section{Methodology}
This section is organized as follows: we first introduce the problem definition, then we demonstrate the utilization of the LID score to differentiate between true-labeled and false-labeled instances. Finally, we present our proposed method, \fullname{}.

\subsection{Problem Definition.}
In this study, we address the problem of training a classification model amidst noisy labels. Let's define the feature space as \(\mathcal{X}\) and \(\mathcal{Y} = \{1, ..., N_c\}\) to be the label space. Our training dataset is represented as \(\tilde{D} = \{(x_i, \tilde{y}_i)\}_{i=1}^{N}\), where each \(\tilde{y}_i = [\tilde{y}_{i,1}, \tilde{y}_{i,2}, ..., \tilde{y}_{i,N_c}]\) is a one-hot vector indicating the \textit{noisy label} for the instance \(x_i\). Here, \(N_c\) denotes the total number of classes. If \(c\) is the noisy label class for \(x_i\), then \(\tilde{y}_{i,j} = 1\) when \(j = c\); otherwise, \(\tilde{y}_{i,j} = 0\). It is crucial to note that a noisy label, \(\tilde{y}_i\), might differ from the actual ground truth label, \(y_i\). An instance is termed a \textit{true-labeled instance} if \(\tilde{y}_i = y_i\), and a \textit{false-labeled instance} if \(\tilde{y}_i \neq y_i\). The set of all features in \(\tilde{D}\) is given by \(X = \{x_i | (x_i, \tilde{y}_i) \in \tilde{D}\}\). Our primary goal is to devise a classification method, denoted as \(f(x;\Theta) \rightarrow \hat{y}\), which can accurately predict the ground-truth label of an instance. In this context, \(\hat{y}_i = [\hat{y}_{i,1}, \hat{y}_{i,2}, ..., \hat{y}_{i,N_c}]\) is a probability distribution over the classes, with \(\sum_{j=1}^{N_c} \hat{y}_{i,j} = 1\).

\input{figures/percentage_distribution}

\subsection{LID and Instance with Noisy Labels.}
\label{ssec:LID}
In this section, we outline the use of a specially designed classifier: \disfullname{} (\disname{}) \(f_{\mathrm{LD}}\), that employs LID as a feature to identify samples with incorrect labels. Prior research  \cite{Ma_Li_Wang_Erfani_Wijewickrema_Schoenebeck_Song_Houle_Bailey_2017} has leveraged the LID scores from the final layer of a trained DNN classifier to characterize adversarial samples. However, in our context, the noise is present in the labels, not in the features. To ensure that \(f_{\mathrm{LD}}\) can detect this noise, we input both the features and label into \(f_{\mathrm{LD}}\). \(f_{\mathrm{LD}}\) consists of three components: a standard backbone model \(g_{\mathrm{LD}}\) (which accepts \(x_i\) as input), a label embedding layer \(g_{\mathrm{LE}}\) that processes the label \(\tilde{y}_i\), and a classification head \(h_{\mathrm{LD}}\) that takes the outputs of \(g_{\mathrm{LD}}\) and \(g_{\mathrm{LE}}\) to produce the final classification. The output from the backbone model \(g_{\mathrm{LD}}(x_i)\) and the label's embedding \(g_{\mathrm{LE}}(\tilde{y}_i)\) are merged as follows:
\begin{equation}
    z(x_i, \tilde{y}_i) = \mathrm{LayerNorm}\left(g_{\mathrm{LD}}(x_i) + g_{\mathrm{LE}}\left(\tilde{y}_i\right)\right)
\end{equation}
The result, \(z(x_i, \tilde{y}_i)\), the \textit{enhanced representation} of $(x_i, \tilde{y}_i)$, is then passed to the classification head \(h_{\mathrm{LD}}\) to predict \(\tilde{y}_i\):
\(
    \hat{y}_{i}^{D} = h_{\mathrm{LD}}\left(z(x_i, \tilde{y}_i)\right)
\).
Here, \(\hat{y}_{i}^{D}\) represents the predicted value of \(\tilde{y}_i\). If we train \(f_{\mathrm{LD}}\) directly using the cross-entropy loss \(\mathcal{L}_{\mathrm{CE}}(\tilde{y}_i, \hat{y}_{i}^{D}) = - \sum_{j=1}^{N_C} \tilde{y}_{i,j} \log(\hat{y}_{i,j}^{D})\), the model will consistently predict \(\tilde{y}_i\). Using the noisy label as the "ground truth" label for measuring prediction accuracy would yield a 100\% accuracy rate. This is because the model's predictions are solely based on the input label \(\tilde{y}_i\). To compel the model to consider both the input features and label, we randomly assign a new label \(\tilde{y}^{*}_i\) for each \(x_i\), ensuring that \(\tilde{y}^{*}_i \neq \tilde{y}_i\). We input the pair \((x_i, \tilde{y}^*_i)\) into \(f_{\mathrm{LD}}\) to obtain another prediction \(\hat{y}_{i}^{*D}\). We then employ the sum of the cross-entropy losses \(\mathcal{L}_{\mathrm{CE}}(\tilde{y}_i, \hat{y}_{i}^{D}) + \mathcal{L}_{\mathrm{CE}}(\tilde{y}_i, \hat{y}_{i}^{*D})\) to train the network. This approach ensures that \(f_{\mathrm{LD}}\) doesn't rely solely on the input label for predictions. 

For LID calculation, we follow the method described in  \cite{ma2018dimensionality} (see Equation 1.5 in supplementary materials). Note that the input to $f_{\mathrm{LD}}$ is $(x_i, \tilde{y}_i)$. Assume that $(x_i, \tilde{y}_i) \in \tilde{D}_B, \tilde{D}_B \subset \tilde{D}$. Here, $\tilde{D}_B$ is the mini-batch drawn from $\tilde{D}$. Let $z(\tilde{D}_B) = \{z(x_i, \tilde{y}_i) | (x_i, \tilde{y}_i) \in \tilde{D}_B\}$. The equation to calculate LID score for $(x_i, \tilde{y}_i)$ can be presented below:
\begin{equation}
\textstyle
    \widehat{\mathrm{LID}}((x_i, \tilde{y}_i), \tilde{D}_B) = -(\frac{1}{k}\sum_{j=1}^k\log\frac{r_{j}(z(x_i, \tilde{y}_i), z(\tilde{D}_B))}{r_{\mathrm{max}}(z(x_i, \tilde{y}_i), z(\tilde{D}_B))})^{-1}.
    \label{eq:LID_DB}
\end{equation}
Here, the term \(r_j(z(x_i, \tilde{y}_i), z(\tilde{D}_B))\) represents the distance of \(z(x_i, \tilde{y}_i)\) to its \(j\)-th nearest neighbor in the set \(\tilde{D}_B\), and \(r_{\mathrm{max}}\) is the neighborhood's radius.
Following the training procedure described above, in order to explore the properties of the LID score of \(z(x_i, y_i)\) in \(f_{\mathrm{LD}}\), we conducted an empirical study on the CIFAR-10 dataset with three types of noise conditions: 20\% instance-dependent noise, 40\% instance-dependent noise, and 50\% symmetric noise. We used ResNet-34  \cite{he2016identity_resnet} as the backbone network \(g_{\mathrm{LD}}\). During the training procedure, we recorded the estimation of the LID score (computed by Equation \ref{eq:LID_DB}) for each instance \((x_i,\tilde{y}_i)\) at every epoch. We then split these LID scores into equal length bins and visualized the percentage distribution of false-labeled instances ($\tilde{y}_i \neq y_i$) and true-labeled instances ($\tilde{y}_i = y_i$) in each bin in Figure \ref{fig:percentage}. As shown in this figure, for all three types of noise conditions, false-labeled instances tend to have higher LID scores compared to true-labeled instances. This observation underscores that, across various noise conditions, LID scores from \disname{} serve as a robust metric to differentiate between true-labeled and false-labeled instances.

\subsection{Proposed Method: \name{}.}
\input{figures/model_architecture}

Informed by these observations, we introduce a collaborative framework, \fullname{}, tailored for learning with noisy labels. The comprehensive structure of \name{} is depicted in Figure \ref{fig:model}, and the pseudo-code of \name{} is presented in the supplementary materials. This framework is bifurcated into two primary subnets: \disname{} \(f_{\mathrm{LD}}\) and \purifullname{} (\puriname{}) \(f_{\mathrm{GE}}\). Specifically, \(f_{\mathrm{GE}}\) operates as a conventional classification model, predicting \(\hat{y}_i\) based on \(x_i\). The training regimen of \name{} unfolds in four distinct phases:

\begin{enumerate}
\setlength{\itemsep}{1pt}
    \item \textbf{Pre-processing:} For an instance \((x_i, \tilde{y}_i)\) drawn from batch \(\tilde{D}_B\), we employ double augmentation to generate two distinct views: \(v^1_i\) and \(v^2_i\). Subsequently, a new label \(\tilde{y}^*_i\) is assigned, ensuring it differs from \(\tilde{y}_i\).
    
    \item \textbf{Prediction and LID Calculation:} Post inputting the features and (features, label) pairs into \disname{} and \puriname{}, predictions are derived from both subnets. Let's denote the predictions from \(f_{\mathrm{GE}}\) as \(\hat{y}^{1,G}_{i}\) and \(\hat{y}^{2,G}_{i}\). Concurrently, \name{} computes the LID scores for both \((v^1_i, \tilde{y}_i)\) and \((v^2_i, \tilde{y}_i)\).
    
    \item \textbf{Loss Weight Assignment:} Utilizing the two LID scores from last step, \name{} allocates weights to each instance. Every instance is endowed with three distinct weights: clean, noisy, hard weights, with each weight catering to a specific loss function.
    
    \item \textbf{Label Update:} \disname{} processes \((v^1_i, \hat{y}^{1,G}_{i})\) and \((v^2_i, \hat{y}^{2,G}_{i})\), deriving LID scores for them. These scores and the difference between prediction from \(f_{\mathrm{LD}}\) and \(f_{\mathrm{GE}}\) subsequently guide the decision on whether to substitute \(\tilde{y}_i\) with a combination of \(\hat{y}^{1,G}_{i}\) and \(\hat{y}^{2,G}_{i}\) for future epochs.
\end{enumerate}

\subsubsection{Pre-processing.}
Consider a mini-batch $\tilde{D}_B = \{(x_i, \tilde{y}_i)\}_{i=1}^{N_B}$ drawn from $\tilde{D}$. This batch can be partitioned into a feature set $X_B = \{x_i | (x_i, \tilde{y}_i) \in \tilde{D}_B \}$ and a label set $\tilde{Y}_B = \{\tilde{y}_i | (x_i, \tilde{y}_i) \in \tilde{D}_B \}$. For each $x_i \in X_B$, \name{} generates two augmented views, $v^1_i$ and $v^2_i$. This two-augmentation design ensures that weight assignment and label update decisions in subsequent steps are not solely dependent on the original input. Such a design can lower the risk of error accumulation during the training procedure. The augmented views lead to:
\(
V^1_B = \{ v^1_i | v^1_i = \mathrm{augmentation1}(x_i), \forall x_i \in X_B \},
V^2_B = \{ v^2_i | v^2_i = \mathrm{augmentation2}(x_i), \forall x_i \in X_B \}.
\)
Same as Section \ref{ssec:LID}, for each label $\tilde{y}_i$ in $\tilde{Y}_B$, a random new label $\tilde{y}^*_i$ is assigned, ensuring that $\tilde{y}^*_i \neq \tilde{y}_i$. 
The resulting set is given by:
\(\tilde{Y}^*_B = \{\tilde{y}^*_i | \tilde{y}^*_i = \mathrm{assignNewLabel}(\tilde{y}_i),\tilde{y}_i \in \tilde{Y}_B \}.\) Consequently, we can define four input pair sets:
\(
    \tilde{D}^{k}_B = \{(v^k_i, \tilde{y}_i) | v^k_i \in V^k_B, \tilde{y}_i \in \tilde{Y}_B\},
    \tilde{D}^{k*}_B = \{(v^k_i, \tilde{y}^*_i) | v^k_i \in V^k_B, \tilde{y}^*_i \in \tilde{Y}^*_B\},
\)
where \( k \in \{1, 2\} \). The \puriname{} subnet $f_{\mathrm{GE}}$ processes $V^1_B$ and $V^2_B$, while the \disname{} subnet $f_{\mathrm{LD}}$ handles $\tilde{D}^{1}_B, \tilde{D}^{2}_B, \tilde{D}^{1*}_B,$ and $\tilde{D}^{2*}_B$.

\subsubsection{Prediction and LID Calculation.}
\label{sssec:pred_LID}
The subnet $f_{\mathrm{GE}}$ takes $V^1_B$ and $V^2_B$ as inputs to predict:
\(
 \hat{Y}^{k,G}_{B} = \{\hat{y}^{k,G}_i | \hat{y}^{k,G}_i = f_{\mathrm{GE}}(v^k_i), v^k_i \in V^k_B \},
\)
where \( k \in \{1, 2\} \).
The subnet $f_{\mathrm{LD}}$ processes $\tilde{D}^{1}_B, \tilde{D}^{2}_B, \tilde{D}^{1*}_B,$ and $\tilde{D}^{2*}_B$ to predict:
\(
\hat{Y}^{k,D}_{B} = \{\hat{y}^{k,D}_i | \hat{y}^{k,D}_i = f_{\mathrm{LD}}(v^k_i, \tilde{y}_i), (v^k_i, \tilde{y}_i) \in \tilde{D}^k_B \}, \\
\hat{Y}^{k*,D}_{B} = \{\hat{y}^{k*,D}_i | \hat{y}^{k*,D}_i = f_{\mathrm{LD}}(v^k_i, \tilde{y}^*_i), (v^k_i, \tilde{y}^*_i) \in \tilde{D}^{k*}_B \},
\)
where \( k \in \{1, 2\} \).  In $f_{\mathrm{LD}}$, each input pair result in an enhanced representations, we use Equation \ref{eq:LID_DB} to calculate LID scores for instances in $\tilde{D}^1_B$ and $\tilde{D}^2_B$:
\begin{gather}
\textstyle
    \widehat{\mathrm{LID}}^{W}(v^k_i, \tilde{y}_i) = \widehat{\mathrm{LID}}((v^k_i, \tilde{y}_i), \tilde{D}^k_B), \\
    \widehat{\mathrm{LID}}^{W}(\tilde{D}^k_B) = \{\widehat{\mathrm{LID}}^{W}\left(v^k_i, \tilde{y}_i\right) | \left(v^k_i, \tilde{y}_i\right) \in \tilde{D}^k_B\},
    \label{eq:LID_weight}
\end{gather}
where \( k \in \{1, 2\} \).
These LID scores are for the weight assignment use only.
After we obtain prediction $\hat{Y}^{1,G}_{B}$ and $\hat{Y}^{2,G}_{B}$ from $f_{\mathrm{GE}}$, we create another two input pair sets:
\(
    \hat{D}^{k}_B = \{(v^k_i, \hat{y}^{k,G}_i) | v^k_i \in V^k_B, \hat{y}^{k,G}_i \in \hat{Y}^{k,G}_B\},
\)
where \( k \in \{1, 2\} \). Both $\hat{D}^{1}_B$ and $\hat{D}^{2}_B$ are fed into the $f_{\mathrm{LD}}$ to obtain enhanced representations. Because we want to compare the LID scores from current noisy label and $f_{\mathrm{GE}}$'s prediction to determine if we want to update the label, we create two union sets, then calculate LID scores within the two sets as follows:
\begin{gather}
\textstyle
    U^k_B = \tilde{D}^{k}_B \cup \hat{D}^{k}_B, \\
    \widehat{\mathrm{LID}}^{U}(v^k_i, \tilde{y}^k_i) = \widehat{\mathrm{LID}}((v^k_i, \tilde{y}^k_i), U^k_B), \\
    \widehat{\mathrm{LID}}^{U}(v^k_i, \hat{y}^{k,G}_i) = \widehat{\mathrm{LID}}((v^k_i, \hat{y}^{k,G}_i), U^k_B), \\
    \begin{split}
    \widehat{\mathrm{LID}}^{U}(U^k_B) &= \{\widehat{\mathrm{LID}}^{U}(v^k_i, \tilde{y}^k_i)| (v^k_i, \tilde{y}^k_i) \in \tilde{D}^{k}_B\} \cup \\
    &\quad \{\widehat{\mathrm{LID}}^{U}(v^k_i, \hat{y}^{k,G}_i)| (v^k_i, \hat{y}^{k,G}_i) \in \hat{D}^{k}_B\},
    \end{split}
\end{gather}\\
where \( k \in \{1, 2\} \). We also collect the output from $f_{\mathrm{LD}}$:
\(
    \hat{Y}^{k,G,D}_{B} = \{\hat{y}^{k,G,D}_i | \hat{y}^{k,G,D}_i = f_{\mathrm{LD}}(v^k_i, \hat{y}^{k,G}_i), (v^k_i, \hat{y}^{k,G}_i) \in \hat{D}^k_B \},
\)
where \( k \in \{1, 2\} \). Note that $\hat{Y}^{1,G,D}_{B}$ and $\hat{Y}^{2,G,D}_{B}$ are only used in label update step and do not participate in loss calculation.

\subsubsection{Loss Weight Assignment.}
\label{sssec:loss}
After estimating the LID scores, we compute weights for each instance. We introduce three types of weights: clean, hard, and noisy. Each type of weight is associated with specific designed loss function. A higher clean weight indicates that the instance is more likely to be a true-labeled instance, while a higher noisy weight suggests the opposite. A high hard weight indicates uncertainty in labeling. As observed in Section \ref{ssec:LID}, instances with smaller LID scores tend to be correctly labeled. Thus, we assign higher clean weights to instances with lower LID scores, higher noisy weights to instances with higher LID scores, and higher hard weights to instances with significant discrepancies in LID scores from two views. To mitigate potential biases in weight assignment, we prefer using \(\widehat{\mathrm{LID}}^{W}\) over \(\widehat{\mathrm{LID}}^{U}\). This preference is due to the observation that the prediction from \(f_{\mathrm{GE}}\) that are identical to the label could lower the label's \(\widehat{\mathrm{LID}}^{U}\) score. Such a decrease does not necessarily indicate the correctness of a label and hence could skew the weight assignment. 
The weights are defined as:

\begin{align}
\textstyle
    q^{k,W}_{\mathrm{low}} &= \mathrm{quantile}(\widehat{\mathrm{LID}}^{W}(\tilde{D}^k_B), \epsilon^{W}_{\mathrm{low}}), \\
    q^{k,W}_{\mathrm{high}} &= \mathrm{quantile}(\widehat{\mathrm{LID}}^{W}(\tilde{D}^k_B), \epsilon^{W}_{\mathrm{high}}), \\
    q^{k,W}_i &= \frac{q^{k,W}_{\mathrm{high}} - \widehat{\mathrm{LID}}^{W}(v^k_i, \tilde{y}_i)}{q^{k,W}_{\mathrm{high}} - q^{k,W}_{\mathrm{low}}}, \\
    w_{i,k} &= \min{(\max{(q^{k,W}_i, 0)}, 1)}, \\
    w_{i,c} &= \min{(w_{i,1}, w_{i,2})}, \\
    w_{i,h} &= |w_{i,1}- w_{i,2}|, \\
    w_{i,n} &= \min{(1 - w_{i,1}, 1 - w_{i,2})},
\end{align}
where \( k \in \{1, 2\} \). Here, \( w_{i,c} \), \( w_{i,h} \), and \( w_{i,n} \) represent clean, hard, and noisy weights, respectively. It's ensured that the sum of \( w_{i,c} \), \( w_{i,h} \), and \( w_{i,n} \) equals 1, which is proved in the supplementary materials. The thresholds \( \epsilon^{W}_{\mathrm{low}} \) and \( \epsilon^{W}_{\mathrm{high}} \) are predefined, satisfying \( 0 \leq \epsilon^{W}_{\mathrm{low}} \leq \epsilon^{W}_{\mathrm{high}} \leq 1 \). Initially, the value of \(\epsilon^{W}_{\mathrm{high}}\) is set low and is then linearly increased over \(\tau\) epochs. This approach ensures that the model does not prematurely allocate a large number of high clean weights, given that the majority of labels have not been refined in the early stages. Only instances with low LID scores are predominantly correctly labeled. For instances with a high clean weight, we employ the cross-entropy loss for optimization. The clean loss is defined as:
\begin{gather}
\textstyle
    \mathcal{L}_{\mathrm{clean, GE}} = w_{i,c} \sum_{k=1}^{2} \mathcal{L}_{\mathrm{CE}}\left(\tilde{y}_i, \hat{y}^{k,G}_i\right), \\
\begin{split}
    \mathcal{L}_{\mathrm{clean, LD}} &= w_{i,c}\sum_{k=1}^{2} \left( \mathcal{L}_{\mathrm{CE}}\left(\tilde{y}_i, \hat{y}^{k,D}_i\right) \right. \\
    &\quad \left. + \lambda^{*}\mathcal{L}_{\mathrm{CE}}\left(\tilde{y}_i, \hat{y}^{k*,D}_i\right) \right).
\end{split}
\end{gather}

Instances with a high \( w_{i,h} \) indicate a significant discrepancy between \( \widehat{\mathrm{LID}}(v^1_i, \tilde{y}_i) \) and \( \widehat{\mathrm{LID}}(v^2_i, \tilde{y}_i) \). This suggests that these instances might be near the decision boundary. While we aim to utilize these instances, the cross-entropy loss is sensitive to label noise. Therefore, we adopt a more robust loss function, the generalized cross entropy (GCE)  \cite{zhang2018generalized}, defined as:
\begin{equation}
\textstyle
    \mathcal{L}_{\mathrm{GCE}}\left(\tilde{y}_i, \hat{y}_i\right) = \sum_{j=1}^{N_C}\tilde{y}_{i,j}\left(1-\left(\hat{y}_{i,j}\right)^q\right)/q,
\end{equation}
where \( q \in (0, 1] \). As shown in  \cite{zhang2018generalized}, this loss function approaches the cross-entropy loss as \( q \rightarrow 0 \) and becomes the MAE loss when \( q=1 \). We set \( q=0.7 \) as recommended by  \cite{zhang2018generalized}. The hard loss is then:
\begin{gather}
\textstyle
    \mathcal{L}_{\mathrm{hard, GE}} = w_{i,h} \sum_{k=1}^{2} \mathcal{L}_{\mathrm{GCE}}\left(\tilde{y}_i, \hat{y}^{k,G}_i\right), \\
\begin{split}
    \mathcal{L}_{\mathrm{hard, LD}} &= w_{i,h} \sum_{k=1}^{2} \left( \mathcal{L}_{\mathrm{GCE}}\left(\tilde{y}_i, \hat{y}^{k,D}_i\right) \right. \\
    &\quad \left. + \lambda^{*}\mathcal{L}_{\mathrm{GCE}}\left(\tilde{y}_i, \hat{y}^{k*,D}_i\right) \right).
\end{split}
\end{gather}

Instances with a high \( w_{i,n} \) are likely to be mislabeled. To leverage these instances without being influenced by label noise, we adopt the CutMix augmentation strategy  \cite{yun2019cutmix}. In essence, CutMix combines two training samples by cutting out a rectangle from one and placing it onto the other \footnote{In this work, we use images as input. While CutMix was designed for images, it hasn't been widely applied to other types of input. For non-image data, other augmentation methods like Mixup \cite{zhang2018mixup} can be considered.}.  For a detailed explanation and methodology of CutMix, readers are referred to  \cite{yun2019cutmix}. We apply CutMix twice for each sample within \( \tilde{D}^1_B \) and \( \tilde{D}^2_B \). The augmented views are defined as:
\begin{align}
    \check{v}^k_{i} &= \mathbf{M}^k v^k_i + (1 - \mathbf{M}^k) v^k_{r_k(i)}, \quad k \in \{1,2\}.\\
    \check{y}^k_{i} &= \lambda_k \tilde{y}_{i} + (1 - \lambda_k) \tilde{y}_{r_k(i)} \quad k \in \{1,2\}.
\end{align}
Here, \( r_1(i) \) and \( r_2(i) \) are random indices for the two views, and \( \mathbf{M}^k \) is a binary mask indicating the regions to combine. The factors \( \lambda_1 \) and \( \lambda_2 \) are sampled from the beta distribution \( Beta(\alpha, \alpha) \), with \( \alpha = 1 \) as suggested by  \cite{yun2019cutmix}. The proportion of the combination is determined by the \( \lambda \) term. Specifically, \( \lambda \) represents the ratio of the original view retained, while \( 1 - \lambda \) denotes the proportion of the other view that's patched in. The CutMix views \( \check{v}^1_{i} \) and \( \check{v}^2_{i} \) are then fed into \( f_{\mathrm{GE}} \) to obtain predictions \( \hat{y}^{1\Check,G}_{i} \) and \( \hat{y}^{2\Check,G}_{i} \). Similarly, \( \left(\check{v}^1_{i}, \check{y}^1_{i}\right) \) and \( \left(\check{v}^2_{i}, \check{y}^2_{i}\right) \) are processed by \( f_{\mathrm{LD}} \) to get \( \hat{y}^{1\Check,D}_{i} \) and \( \hat{y}^{2\Check,D}_{i} \).
The loss for CutMix instances is defined as:
\begin{align}
\begin{split}
    \mathcal{L}^\prime_{\mathrm{noisy, GE}} &= \sum_{k=1}^{2} \left[ \lambda_k \mathcal{L}_{\mathrm{CE}}\left(\tilde{y}_{i}, \hat{y}^{k\Check,G}_{i}\right) \right. \\
    &\quad \left. + \left(1 - \lambda_k\right) \mathcal{L}_{\mathrm{CE}}\left(\tilde{y}_{r_k(i)}, \hat{y}^{k\Check,G}\right) \right],
\end{split}\\
\begin{split}
    \mathcal{L}^\prime_{\mathrm{noisy, LD}} &= \sum_{k=1}^{2} \left[ \lambda_k \mathcal{L}_{\mathrm{CE}}\left(\tilde{y}_{i}, \hat{y}^{k\Check,D}_{i}\right) \right. \\
    &\quad \left. + \left(1 - \lambda_k\right) \mathcal{L}_{\mathrm{CE}}\left(\tilde{y}_{r_k(i)}, \hat{y}^{k\Check,D}\right) \right].
\end{split}
\end{align}
To enhance the learning from instances with high noise weights in \(f_{\mathrm{LD}}\), we also employ a consistency loss. This loss, based on cosine similarity, ensures consistent predictions between \(\tilde{y}_i\) and \(\tilde{y}^*_i\) without relying on label guidance. The consistency loss for \(f_{\mathrm{LD}}\) is:
\begin{equation}
\textstyle
    \mathcal{L}_{\mathrm{cons, LD}} = \sum_{k=1}^{2}\left(1 - \cos{\left(\hat{y}^{k,D}_i, \hat{y}^{k*,D}_i\right)}\right).
\end{equation}
We combine the consistency loss and CutMix loss to get the noisy loss as:
\begin{align}
    \mathcal{L}_{\mathrm{noisy, GE}} &= w_{i,n} \mathcal{L}^\prime_{\mathrm{noisy, GE}}, \\
    \mathcal{L}_{\mathrm{noisy, LD}} &= w_{i,n} (\mathcal{L}^\prime_{\mathrm{noisy, LD}} + \lambda_{\mathrm{cons}} \mathcal{L}_{\mathrm{cons, LD}}).
\end{align}

The overall training objectives for \(f_{\mathrm{GE}}\) and \(f_{\mathrm{LD}}\) combine the clean, hard, and noisy losses:
\begin{align}
    \mathcal{L}_{\mathrm{GE}} &= \mathcal{L}_{\mathrm{clean, GE}} + \mathcal{L}_{\mathrm{hard, GE}} + \mathcal{L}_{\mathrm{noisy, GE}}, \\
    \mathcal{L}_{\mathrm{LD}} &= \mathcal{L}_{\mathrm{clean, LD}} + \mathcal{L}_{\mathrm{hard, LD}} + \mathcal{L}_{\mathrm{noisy, LD}}.
\end{align}
Both \(f_{\mathrm{GE}}\) and \(f_{\mathrm{LD}}\) are optimized separately using their respective loss functions.

\subsubsection{Label Update.}
For determining whether to update the label based on the prediction from \(f_{\mathrm{GE}}\), we consider both the LID scores and the prediction difference between \(f_{\mathrm{GE}}\) and \(f_{\mathrm{LD}}\). As discussed in Section \ref{ssec:LID}, instances with smaller LID scores are more likely to be correctly labeled. If the LID scores associated with \(f_{\mathrm{GE}}\)'s prediction are smaller than the current label's scores, then the prediction is more likely to be accurate. The prediction difference is defined as:
\begin{align}
    \Delta\tilde{y}^k_i &= {\textstyle\sum}_{j=1}^{N_c}|\hat{y}^{k,G}_{i,j} - \hat{y}^{k,D}_{i,j}|, \quad k \in \{1,2\}.\\
    \Delta\hat{y}^k_i &= {\textstyle\sum}_{j=1}^{N_c}|\hat{y}^{k,G}_{i,j} - \hat{y}^{k,G, D}_{i,j}|, \quad k \in \{1,2\}.
\end{align}
The principle of agreement maximization suggests that different models are less likely to agree on incorrect labels \cite{wei2020combating}. The \(\Delta\) value measures the level of disagreement between \(f_{\mathrm{GE}}\) and \(f_{\mathrm{LD}}\). A larger \(\Delta\) value indicates that the corresponding prediction or label is less likely to be correct. Generally, if a prediction has a smaller LID score and a smaller \(\Delta\) compared to the current label, it's a candidate for label replacement. Using the LID scores computed in Section \ref{sssec:pred_LID}, \name{} make decision on label updating as follows:
\begin{gather}
\textstyle
    q^{k,U}_{\mathrm{low}} = \mathrm{quantile}(\widehat{\mathrm{LID}}^{U}(U^k_B), \epsilon^{U}_{\mathrm{low}}), \\
    q^{k,U}_{\mathrm{high}} = \mathrm{quantile}(\widehat{\mathrm{LID}}^{U}(U^k_B), \epsilon^{U}_{\mathrm{high}}), \\
    \tilde{q}^{k,U}_i = (q^{k,U}_{\mathrm{high}} - \widehat{\mathrm{LID}}^{U}(v^k_i, \tilde{y}_i))/(q^{k,U}_{\mathrm{high}} - q^{k,U}_{\mathrm{low}}), \\
    \hat{q}^{k,U}_i = (q^{k,U}_{\mathrm{high}} - \widehat{\mathrm{LID}}^{U}(v^k_i, \hat{y}^{k,G}_i))/(q^{k,U}_{\mathrm{high}} - q^{k,U}_{\mathrm{low}}), \\
    \tilde{t}^{k}_i = \min{(1,\max{(0,\tilde{q}^{k,U}_i * (2 - \Delta\tilde{y}^k_i)/2)})}, \\
    \hat{t}^{k}_i = \min{(1,\max{(0,\hat{q}^{k,U}_i * (2 - \Delta\hat{y}^k_i)/2)})},
\end{gather}
where \(k \in \{1,2\}\), $\epsilon^{U}_{\mathrm{low}}$ and $\epsilon^{U}_{\mathrm{high}}$ are thresholds. Mirroring the approach of \(\epsilon^{W}_{\mathrm{high}}\), \(\epsilon^{U}_{\mathrm{high}}\) starts low and linearly rises over \(\tau\) epochs, enabling the model to judiciously assess the reliability of labels and predictions. The \(\Delta\) values are normalized to the [0,1] range using \((2 - \Delta)/2\), which is elaborated in the supplementary materials.  In these equations, predictions or labels with smaller LID values and smaller cross-subnet differences have larger \(t\) values, and vice versa. The process of converting both $\hat{y}^{1,G}_i$ and $\hat{y}^{2,G}_i$ to one-hot label vectors is as follows:
\begin{equation}
\textstyle
    \acute{y}_i =  [\acute{y}_{i,0}, \acute{y}_{i,1}, ...\acute{y}_{i, N_c}]
\end{equation}
\[\textstyle\acute{y}_{i,l} = 
\begin{cases}
1, & \text{if } l = \underset{j}{\mathrm{argmax}}(\hat{y}_{i,j}) \\
0, & \text{otherwise} 
\end{cases} \]
We determine whether to update the label as follows:
\begin{align}
\textstyle
    \begin{split}
        \text{cond} &= \hat{t}^{1}_i > \tilde{t}^{1}_i \And{\hat{t}^{2}_i > \tilde{t}^{2}_i} \\
        &\quad \And{\hat{t}^{1}_i > \epsilon_{k}} \And{\hat{t}^{2}_i > \epsilon_k} \And{\acute{y}^{1,G}_i = \acute{y}^{2,G}_i}
    \label{eq:cond}
    \end{split}\\
    \tilde{y}_i &= 
    \begin{cases}
        \acute{y}^{1,G}_i, & \text{if cond}  \\
        \tilde{y}_i, & \text{otherwise}
    \end{cases}
\end{align}
In this decision-making process, a label is only updated when the prediction's \(t\) value from both views surpasses a predefined threshold \(\epsilon_k\) and is higher than higher than the \(t\) value of the corresponding label. Both predictions \(\acute{y}^{1,G}_i\) and \(\acute{y}^{2,G}_{i}\) must also be equal, minimizing the chance of assigning an incorrect label to instance \(x_i\). Note that the new $\tilde{y}_i$ is used for the next epoch,  ensuring that in current epoch, the loss calculation in Section \ref{sssec:loss} is not affected by the label update.

%% file: figures/percentage_distribution.tex
\begin{figure*}[th!]
\begin{subfigure}[t]{\textwidth}
    \centering
    \includegraphics[width=0.9\textwidth]{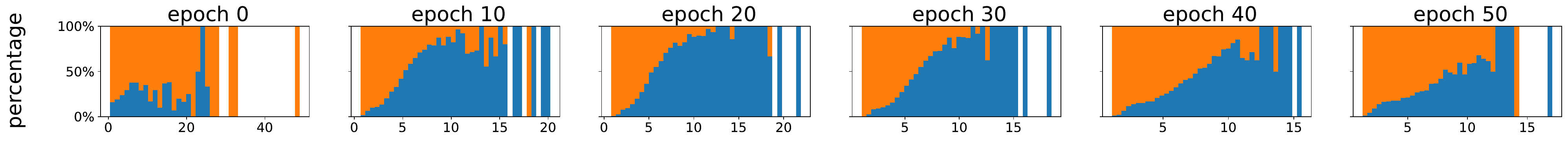}
\end{subfigure}
\begin{subfigure}[t]{\textwidth}
    \centering
    \includegraphics[width=0.9\textwidth]{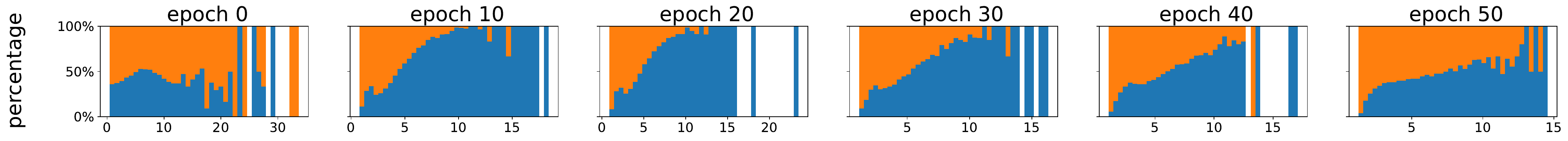}
\end{subfigure}
\begin{subfigure}[t]{\textwidth}
    \centering
    \includegraphics[width=0.9\textwidth]{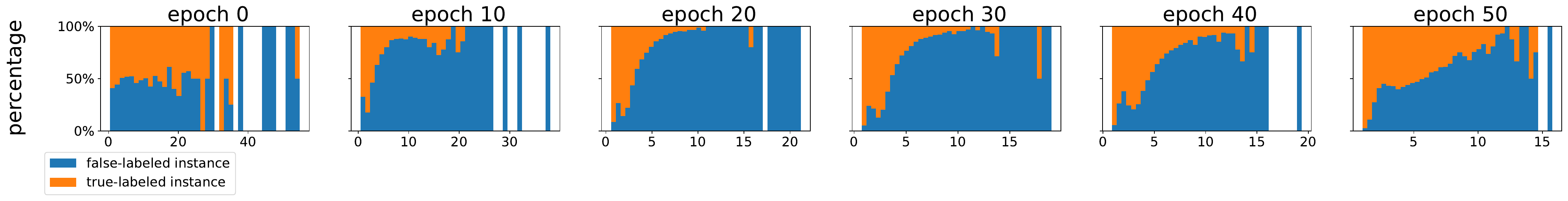}
\end{subfigure}
\caption{Distribution of LID scores for true-labeled versus false-labeled instances in three noise conditions. The heights of the orange and blue bars represent the proportions of true-labeled and false-labeled instances' LID scores within specific bins, respectively. LID scores are based on the enhanced representation of features and labels in the \disname{}. From top to bottom, the noise conditions for the three figures are: 20\% instance-dependent noise, 40\% instance-dependent noise, and 50\% symmetric noise.}
\label{fig:percentage}
\end{figure*}

%% file: figures/model_architecture.tex
\begin{figure*}[th]
    \centering
    \includegraphics[width=0.7\textwidth]{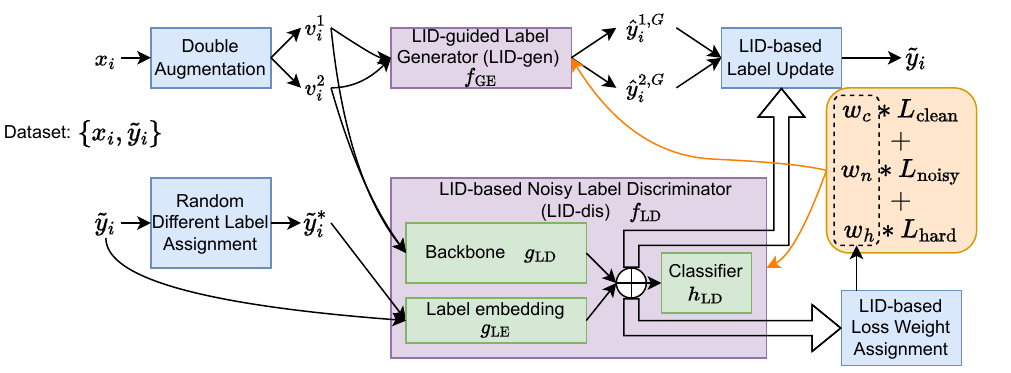}
    \caption{The overall framework of \name{}.}
    \label{fig:model}
\end{figure*}

%% file: sections/sec4_experiment.tex
\section{Experiments}
In this section, we assess the performance of \name{} across various noise settings. 
We also present ablation studies to validate the contribution of each component.

{\bf Experiment Setup.} \name{} is evaluated on CIFAR-10 \cite{krizhevsky2009learning} with three type of noise: symmetric (sym.), asymmetric (asym.) and instance-dependent (inst.) noise. Sym.\@ noise involves uniformly flipping labels at random, while asym.\@ noise flips labels between neighboring classes at a fixed probability, following methods in \cite{han2018co, li2019dividemix}. Inst.\@ noise is generated per instance using a truncated Gaussian distribution as per \cite{li2023disc, xia2020part}. Noise ratios are set at \{20\%, 50\%, 80\%\} for sym.\@ noise, \(40\%\) for asym.\@ noise, and \{20\%, 40\%, 60\%\} for inst.\@ noise, aligning with settings in \cite{Zhu_Liu_Liu_2021_GAL, li2023disc}. Additionally, experiments are conducted on CIFAR-10N \cite{wei2021learning}, a real-world noisy dataset with re-annotated CIFAR-10 images. CIFAR-10N provides three submitted labels (i.e., Random 1, 2, 3) per image, aggregated to create an Aggregate and a Worst label. The Aggregate, Random 1, and Worst label are used in experiment. ResNet-18 \cite{he2016identity_resnet} serves as the backbone for CIFAR-10 with sym.\@ and asym.\@ noise, while ResNet-34 is used for CIFAR-10 with inst.\@ noise and CIFAR-10N. Additional details are provided in the supplementary materials.

\input{tables/performance_cifar10_dual}
\input{tables/performance_cifar10_real}


\subsection{Experiment Results.}
Table \ref{tab:noise_10_sym} shows that \name{} consistently ranks among the top three in prediction accuracy on the CIFAR-10 dataset under both sym.\@ and asym.\@ noise conditions, achieving the highest average accuracy across four scenarios. Its robustness to various noise ratios and types stands out. In Table \ref{tab:noise_10_ins}, \name{} achieves the highest average accuracy under inst.\@ noise, notably excelling at an 80\% noise ratio. Table \ref{tab:noise_10_real} further demonstrates \name{}'s superior performance and robustness under real-world noise settings, particularly under high noise conditions (Worst, 40\% noise). These findings underscore the robustness and superior generalization capability of \name{}.


\input{tables/performance_ablation}
\subsection{Ablation Study.}
In our ablation study, we examine four \name{} variants to assess the impact of the dual-view design and the three loss types. Table \ref{tab:ablation} presents these variants: the first uses only the original features for weight assignment and label updates; the second to fourth exclude both noise \& hard loss, noise loss only, or,
hard loss only, respectively. All these variants still use LID scores to assign weight and determine label updates.
%
%
Results show that the complete \name{} model outperforms its variants. The performance gap between the single-view variant and \name{} highlights the dual-view approach's role in reducing errors and enhancing model robustness. The lesser performance of the latter three variants compared to \name{} confirms that combining all three loss types effectively utilizes information from both correctly and incorrectly labeled instances.

%% file: tables/performance_cifar10_dual.tex
\begin{table*}[!ht]
\centering
\caption{Comparative performance on CIFAR-10 under different noise settings. Our implementations are marked with *; others are from \cite{li2023disc}. {\bf Bold} scores are the highest, and {\underline{underlined}} scores the second highest in each setting.
}
\noindent
\begin{subtable}[t]{.55\textwidth}
\caption{CIFAR-10, sym.\@ and asym.\@ noise}
\centering
\resizebox{\textwidth}{!}{
\input{tables/cifar10_sym_tab}
}
\label{tab:noise_10_sym}
\end{subtable}%
\begin{subtable}[t]{.45\textwidth}
\caption{CIFAR-10, inst.\@ noise}
\centering
\resizebox{\textwidth}{!}{
\input{tables/cifar10_ins_tab}
}
\captionlistentry{}
\label{tab:noise_10_ins}
\end{subtable}    

\end{table*}


%% file: tables/cifar10_sym_tab.tex
\begin{tabular}[t]{lccccc}
\toprule
Methods & Sym.\@ 20\% & Sym.\@ 50\% & Sym.\@ 80\% & Asym.\@ 40\% & Average \\
\midrule
Cross Entropy & 83.31 ± 0.09 & 56.41 ± 0.32 & 18.52 ± 0.16 & 77.06 ± 0.26 & 58.83 \\
Mixup \cite{zhang2018mixup} & 90.17 ± 0.12 & 70.94 ± 0.26 & 47.15 ± 0.37 & 82.68 ± 0.38 & 72.74 \\
Decoupling \cite{malach2017decoupling} & 85.40 ± 0.12 & 68.57 ± 0.34 & 41.08 ± 0.24 & 78.67 ± 0.81 & 68.43 \\
Co-teaching \cite{han2018co}& 87.95 ± 0.07 & 48.60 ± 0.19 & 17.48 ± 0.11 & 71.14 ± 0.32 & 56.29 \\
JointOptim \cite{tanaka2018joint} & 91.34 ± 0.40 & 89.28 ± 0.74 & 59.67 ± 0.27 & 90.63 ± 0.39 & 82.73 \\
Co-teaching+\cite{yu2019does} & 87.20 ± 0.08 & 54.24 ± 0.23 & 22.26 ± 0.55 & 79.91 ± 0.46 & 60.90 \\
GCE \cite{zhang2018generalized} & 90.05 ± 0.10 & 79.40 ± 0.20 & 20.67 ± 0.11 & 74.73 ± 0.39 & 66.21 \\
PENCIL \cite{yi2019probabilistic} & 88.02 ± 0.90 & 70.44 ± 1.09 & 23.20 ± 0.81 & 76.91 ± 0.26 & 64.64 \\
JoCoR \cite{wei2020combating}& 89.46 ± 0.04 & 54.33 ± 0.12 & 18.31 ± 0.11 & 70.98 ± 0.21 & 58.27 \\
DivideMix* \cite{li2019dividemix} & 92.87 ± 0.46 & \underline{94.75 ± 0.14} & 81.25 ± 0.26 & 91.88 ± 0.12 & 90.19 \\
ELR \cite{liu2020early} & 90.35 ± 0.04 & 87.40 ± 3.86 & 55.69 ± 1.00 & 89.77 ± 0.12 & 80.80 \\
ELR+ \cite{liu2020early} & 95.27 ± 0.11 & 94.41 ± 0.11 & \underline{81.86 ± 0.23} & 91.38 ± 0.50 & 90.73 \\
Co-learning \cite{Tan_Xia_Wu_Li_2021_colearning} & 92.14 ± 0.09 & 77.99 ± 0.65 & 43.80 ± 0.76 & 82.70 ± 0.40 & 74.16 \\
GJS* \cite{englesson2021generalized} & 83.57 ± 1.24 & 50.26 ± 2.54 & 15.49 ± 0.18 & 85.64 ± 1.37 & 58.74 \\
DISC* \cite{li2023disc} & \textbf{95.99 ± 0.15} & \textbf{95.03 ± 0.12} & 81.84 ± 0.21 & \underline{94.20 ± 0.07} & \underline{91.69} \\
{\it \name{} (ours)} & \underline{95.32 ± 0.08} & 93.64 ± 0.11 & \textbf{84.42 ± 0.20} & \textbf{94.67 ± 0.11} & \textbf{92.01} \\
\bottomrule
\end{tabular}

%% file: tables/cifar10_ins_tab.tex
\begin{tabular}[t]{lcccc}
\toprule
Methods & Inst.\@ 20\% & Inst.\@ 40\% & Inst.\@ 60\% & Average \\
\midrule
Cross Entropy & 83.93 ± 0.15 & 67.64 ± 0.26 & 43.83 ± 0.33 & 65.13 \\
Forward T \cite{patrini2017making_DMI} & 87.22 ± 1.60 & 79.37 ± 2.72 & 66.56 ± 4.90 & 77.72 \\
DMI \cite{xu2019l_dmi}& 88.57 ± 0.60 & 82.82 ± 1.49 & 69.94 ± 1.34 & 80.44 \\
Mixup \cite{zhang2018mixup} & 87.71 ± 0.66 & 82.65 ± 0.38 & 58.59 ± 0.58 & 76.32 \\
GCE  \cite{zhang2018generalized} & 89.80 ± 0.12 & 78.95 ± 0.15 & 60.76 ± 3.08 & 76.50 \\
Co-teaching \cite{han2018co} & 88.87 ± 0.24 & 73.00 ± 1.24 & 62.51 ± 1.98 & 74.79 \\
Co-teaching+ \cite{yu2019does} & 89.80 ± 0.28 & 73.78 ± 1.39 & 59.22 ± 6.34 & 74.27 \\
JoCoR \cite{wei2020combating} & 88.78 ± 0.15 & 71.64 ± 3.09 & 63.46 ± 1.58 & 74.63 \\
Reweight-R \cite{xia2019anchor} & 90.04 ± 0.46 & 84.11 ± 2.47 & 72.18 ± 2.47 & 82.11 \\
Peer Loss \cite{liu2020peer} & 89.12 ± 0.76 & 83.26 ± 0.42 & 74.53 ± 1.22 & 82.30 \\
DivideMix* \cite{li2019dividemix} & 92.95 ± 0.29 & \underline{94.99 ± 0.14} & 89.30 ± 1.32 & 92.41 \\
CORSES\textsuperscript{2} \cite{cheng2021learning} & 91.14 ± 0.46 & 83.67 ± 1.29 & 77.68 ± 2.24 & 84.16 \\
CAL \cite{Zhu_Liu_Liu_2021_GAL}& 92.01 ± 0.12 & 84.96 ± 1.25 & 79.82 ± 2.56 & 85.60 \\
DISC* \cite{li2023disc} & \textbf{96.34 ± 0.13} & \textbf{95.27 ± 0.21} & \underline{91.15 ± 2.20} & 94.25 \\
{\it \name{} (ours)} & \underline{95.73 ± 0.10} & 94.66 ± 0.11 & \textbf{92.45 ± 1.25} & \textbf{94.28} \\
\bottomrule
\end{tabular}%

%% file: tables/performance_cifar10_real.tex
\begin{table}[!ht]
\centering
\caption{Performance comparison on CIFAR-10N. Our implementations are marked with *; others are from \cite{liu2022robust}.}
\resizebox{\columnwidth}{!}{%
\input{tables/cifar10_real_tab}
}
\label{tab:noise_10_real}
\end{table}

%% file: tables/cifar10_real_tab.tex
\begin{tabular}{lcccc}
\toprule
Methods & Aggregate & Random & Worst & Average \\
\midrule
Cross Entropy & 87.77 ± 0.38 & 85.02 ± 0.65 & 77.69 ± 1.55 & 83.49 \\
Forward T \cite{patrini2017making_DMI} & 88.24 ± 0.22 & 86.88 ± 0.50 & 79.79 ± 0.46 & 84.97 \\
Co-teaching \cite{han2018co} & 91.20 ± 0.13 & 90.33 ± 0.13 & 83.83 ± 0.13 & 88.45 \\
ELR+ \cite{liu2020early}& 94.83 ± 0.10 & 94.43 ± 0.41 & 91.09 ± 1.60 & 93.45 \\
CORES\textsuperscript{2} \cite{cheng2021learning}& 95.25 ± 0.09 & 94.45 ± 0.14 & \underline{91.66 ± 0.09} & 93.79 \\
DISC* \cite{li2023disc} & \textbf{95.96 ± 0.04} & \textbf{95.33 ± 0.12} & 90.20 ± 0.24 & \underline{93.83} \\
{\it \name{} (ours)} & \underline{95.74 ± 0.14} & \underline{95.21 ± 0.27} & \textbf{92.65 ± 0.10} & \textbf{94.53} \\
\bottomrule
\end{tabular}%

%% file: tables/performance_ablation.tex
\begin{table}[!htb]
    \centering
    \caption{Ablation study for two views and loss types.}
    \resizebox{0.85\columnwidth}{!}{%
    \begin{tabular}{cc}
    \toprule
    Variations	& CIFAR-10, 40\% Inst.\@ Noise \\
    \midrule
    \name{} w/o two views	& 84.31 ± 0.59 \\
        \name{} w/o noise and hard loss	        & 91.06 ± 0.22 \\
    \name{} w/o noise loss	        & 91.36 ± 0.34 \\
    \name{} w/o hard loss	        & 93.56 ± 0.25 \\
    \name{}	                        &   \textbf{94.66 ± 0.11} \\
    \bottomrule
    \end{tabular}%
    }
    \label{tab:ablation}
\end{table}


%% file: sections/sec5_conclusion.tex
\section{Conclusion}

In this study, we present \name{}, a novel framework designed for learning with noisy labels.
It is composed of two key subnets: \disfullname{} (\disname{}) and \purifullname{} (\puriname{}). Both two subnets leverage two augmented views of features for each instance. The \disname{} assimilates features and labels of training samples to create enhanced representations. \name{} employs LID scores from enhanced representations to weight the loss function for both subnets.
\puriname{} suggests pseudo-labels, and \disname{} process pseudo-labels along with two views to derive LID scores. These LID scores and the discrepancies in prediction from the two subnets inform the label correction decisions. This dual-view and dual-subnet approach significantly reduces the risk of errors and enhances the overall effectiveness of the framework. After training, \puriname{} is ready to be deployed as the classifier. Extensive evaluations demonstrate \name{}'s superiority over existing state of the arts in various noise settings, notably improving prediction accuracy.

%% file: sections/sec6_acknowledgement.tex
\section*{Acknowledgments}
This work was supported by the AFRI award no. 2020-67021-32459 from USDA NIFA. We also thank the WPI DAISY group for their input on this research.

%% file: sections/supp.tex
\appendix
\section{Local Intrinsic Dimensionality (LID)}
Local Intrinsic Dimensionality (LID) serves as an expansion-centric metric, capturing the intrinsic dimensionality of a data's underlying subspace or submanifold \cite{houle2017local}. Within intrinsic dimensionality theory, expansion models quantify the growth rate of in the number of data objects encountered as the distance from a reference sample expands \cite{Ma_Li_Wang_Erfani_Wijewickrema_Schoenebeck_Song_Houle_Bailey_2017}. 
To provide an intuitive perspective, consider a Euclidean space where the volume of an $m$-dimensional ball scales in proportion to $r^m$ as its size is adjusted by a factor of $r$. Given this relationship between volume growth and distance, dimension $m$ can be inferred using:
\begin{equation}
    \frac{V_2}{V_1} = (\frac{r_2}{r_1})^m \Rightarrow m = \frac{\ln{(V_2/V_1)}}{\ln{(r_2/r_1)}}
    \label{eq:es_dim}
\end{equation}
By interpreting the probability distribution as a volume surrogate, traditional expansion models offer a local perspective on data's dimensional structure, as their estimates are confined to the vicinity of the sample of interest. Adapting the expansion dimension concept to the statistical realm of continuous distance distributions results in LID's formal definition \cite{houle2017local}:

\textbf{Definition 1} (Local Intrinsic Dimensionality).
\textit{For a data sample $x \in X$, let $r > 0$ represent the distance from $x$ to its neighboring data samples. If the cumulative distribution function $F(r)$ is both positive and continuously differentiable at a distance $r > 0$, then the LID of $x$ at distance $r$ is expressed as:
\begin{equation}
    \mathrm{LID}_F(r) \triangleq \lim_{\epsilon \rightarrow 0}\frac{\ln{(F((1+\epsilon)r)/F(r))}}{\ln{(1+\epsilon)}} = \frac{rF^\prime(r)}{F(r)}
    \label{eq:LID_Fr}
\end{equation}
whenever the limit exists. The LID at $x$ is  subsequently defined as the limit as radius $r \rightarrow 0$:}
\begin{equation}
    \mathrm{LID}_F = \lim_{r \rightarrow 0}\mathrm{LID}_F(r)
    \label{eq:LID_F}
\end{equation}

\textbf{Estimation of LID.}
Consider a reference sample point $x \sim \mathcal{X}$, where $\mathcal{X}$ denotes a global data distribution. Each sample $x_{*} \sim \mathcal{X}$ being associated with the distance value $d(x, x_{*})$ relative to $x$. When examining a dataset $X$ derived from $\mathcal{X}$, the smallest $k$ nearest neighbor distances from $x$ can be interpreted as extreme events tied to the lower end of the induced distance distribution\cite{ma2018dimensionality}. Delving into the statistical theory of extreme values, it becomes evident that the tails of continuous distance distributions tend to align with the Generalized Pareto Distribution (GPD), a type of power-law distribution\cite{Coles_2001}. In this work, we  adopt the methodology from \cite{ma2018dimensionality}, and employ the Maximum Likelihood Estimator, represented as:
\begin{equation}
    \widehat{\mathrm{LID}}(x) = -\left(\frac{1}{k}\sum_{i=1}^k\log\frac{r_i(x)}{r_{\mathrm{max}}(x)}\right)^{-1}
    \label{eq:LID_E}
\end{equation}
Here, $r_{i}(x)$ signifies the distance between $x$ and its $i$-th nearest neighbor, while $r_{\mathrm{max}}(x)$ represents the maximum of these neighbor distances. It's crucial to understand that the $\mathrm{LID}$ defined in \eqref{eq:LID_F} is a \textit{distributional} quantity, and the $\widehat{\mathrm{LID}}$ defined in \eqref{eq:LID_E}  serves as its \textit{estimate}.

However, in practice, computing neighborhoods with respect to the entire feature set $X$ can be prohibitively expensive, we will estimate LID of a training example $x$ from its $k$-nearest neighbor set within a batch randomly selected from $X$. Consider a L-layer neural network $h: \mathcal{X} \rightarrow \mathbb{R}^c$, where \(h_j\) is the transformation at the \(j\)-th layer, and given a batch \(X_B \subset X\) and a reference point \(x\), the LID score of \(x\) is estimated as\cite{ma2018dimensionality}:
\begin{equation}
    \widehat{\mathrm{LID}}(x, X_B) = -\left(\frac{1}{k}\sum_{i=1}^k\log\frac{r_i(h_j(x), h_j(X_B))}{r_{\mathrm{max}}(h_j(x), h_j(X_B))}\right)^{-1}
    \label{eq:LID_EB}
\end{equation}
In this equation, \(h_j(x)\) is the output from the \(j\)-th layer of the network. The term \(r_i(h_j(x), h_j(X_B))\) represents the distance of \(h_j(x)\) to its \(i\)-th nearest neighbor in the transformed set \(h_j(X_B)\), and \(r_{\mathrm{max}}\) is the neighborhood's radius. The value \(\widehat{\mathrm{LID}}(x, X_B)\) indicates the dimensional complexity of the local subspace surrounding \(x\) after the transformation by \(h_j\). If the batch is adequately large, ensuring the \(k\)-nearest neighbor sets remain in the vicinity of $h_j(x)$, the estimate of LID at $h_j(x)$ within the batch serves as an approximation to the value that would have been computed within the full dataset $h_j(X)$.

\begin{algorithm*}
\caption{\name{} algorithm.}
\label{alg:colafier}
\begin{algorithmic}
    \STATE \textbf{Input}: noisy dataset \(\tilde{D} = \{(x_i, \tilde{y}_i)\}\), start epoch \(T_{0}\), total epochs \(T_{\max}\), total number of batches \(B_{\max}\), \disname{} \({f_{\mathrm{LD}}(\Theta_{\mathrm{LD}})}\), \puriname{}  \(f_{\mathrm{GE}}(\Theta_{\mathrm{GE}})\), \(\lambda^{*}, \lambda_{cons}, \epsilon^{W}_{\mathrm{low}}, \epsilon^{W}_{\mathrm{high}}, \epsilon^{U}_{\mathrm{low}}, \epsilon^{U}_{\mathrm{high}}, \tau, \epsilon_{k}\).
    \STATE \textbf{Output}: \puriname{}  \(f_{\mathrm{GE}}(\Theta_{\mathrm{GE}})\)
    \FOR{\(T = 1, ..., T_{\max}\)}
    \FOR{\(B = 1, ..., B_{\max}\)}
    \STATE obtain a mini-batch \(\tilde{D}_B = \{(x_i, \tilde{y}_i)\}_{i=1}^{N_{B}}\)
    \STATE obtain view sets \(V^1_B\) and \(V^2_B\), and input pair sets \(\tilde{D}^{1}_B, \tilde{D}^{2}_B, \tilde{D}^{1*}_B, \tilde{D}^{2*}_B\)
    \STATE obtain prediction sets: \(\hat{Y}^{k,G}_B\) from \(f_{\mathrm{GE}}\), and \(\hat{Y}^{k,D}_B, \hat{Y}^{k*,D}_B\) from \(f_{\mathrm{LD}}\), where \(k \in \{1,2\}\) 

    \IF{\(T \leq T_{0}\)}
        \STATE obtain \(\mathcal{L}_{\mathrm{GE}} = \sum_{k=1}^2 \left(\mathcal{L}_{\mathrm{CE}}(\tilde{y}_i, \hat{y}^{k,G}_i)\right)\), \(\mathcal{L}_{\mathrm{GE}} = \sum_{k=1}^2 \left( \mathcal{L}_{\mathrm{CE}}(\tilde{y}_i, \hat{y}^{k,D}_i) + \lambda^{*} \mathcal{L}_{\mathrm{CE}}(\tilde{y}_i, \hat{y}^{k*,D}_i) \right)\) \COMMENT{Warm-up}
    \ELSE
        \STATE obtain \(\widehat{\mathrm{LID}}^{W}(\tilde{D}^1_B)\), and \(\widehat{\mathrm{LID}}^{W}(\tilde{D}^2_B)\) \COMMENT{Using Equation 3.3-3.4 to get LID scores for weight assignment}
        \STATE obtain \(\hat{D}^1_B, \hat{D}^2_B\) \COMMENT{Input pairs for predictions from $f_{\mathrm{GE}}$}
        \STATE obtain \(\hat{U}^k_B = \tilde{D}^k_B \cup \hat{D}^k_B\), and \(\widehat{\mathrm{LID}}^{U}(\hat{U}^k_B)\) \COMMENT{Using Equation 3.5-3.8 to get LID scores for label update}
        \STATE obtain \(\{w_{i,c}\}, \{w_{i,h}\}\), and \(\{w_{i,n}\}\) \COMMENT{Using Equation 3.9 - 3.15 to get weights for each loss term}
        \STATE obtain \(\mathcal{L}_{\mathrm{clean, GE}}, \mathcal{L}_{\mathrm{hard, GE}}, \mathcal{L}_{\mathrm{noisy, GE}}, \mathcal{L}_{\mathrm{clean, LD}}, \mathcal{L}_{\mathrm{hard, LD}}, \mathcal{L}_{\mathrm{noisy, LD}}\) \COMMENT{Using Equation 3.16 - 3.27 to calculate weighted clean, hard, and noisy loss}
        \STATE obtain \(\mathcal{L}_{\mathrm{GE}} = \mathcal{L}_{\mathrm{clean, GE}} + \mathcal{L}_{\mathrm{hard, GE}} + \mathcal{L}_{\mathrm{noisy, GE}}, \mathcal{L}_{\mathrm{LD}} = \mathcal{L}_{\mathrm{clean, LD}} + \mathcal{L}_{\mathrm{hard, LD}} + \mathcal{L}_{\mathrm{noisy, LD}}\)
    \ENDIF
    \STATE \(\Theta^{B+1}_{\mathrm{GE}} = \mathrm{AdamW}(\mathcal{L}_{\mathrm{GE}}, \Theta^{B}_{\mathrm{GE}})\), and \(\Theta^{B+1}_{\mathrm{LD}} = \mathrm{AdamW}(\mathcal{L}_{\mathrm{LD}}, \Theta^{B}_{\mathrm{LD}})\)
    \IF{\(T > T_{0}\)}
        \FOR{\(i = 1, ..., N_{B}\)}
        \STATE obtain \(\Delta\tilde{y}^k_i, \Delta\hat{y}^k_i\), where \(k \in {1,2}\) \COMMENT{Using Equation 3.30 and 3.31 to calculate prediction difference}
        \STATE obtain \(\tilde{t}^k_i, \hat{t}^k_i, \acute{y}^{k,G}_i\), where \(k \in {1,2}\), then determine whether to update label \(\tilde{y}_i\) with \(\acute{y}^{k,G}_i\) or not \COMMENT{Using Equation 3.32-3.40 to make decision on label update}
        \ENDFOR
    \ENDIF
    \ENDFOR
    \ENDFOR
\end{algorithmic}
\end{algorithm*}

\section{The Pseudo Code of \name{}}
The pseudo-code for \name{} is presented in Algorithm \ref{alg:colafier}. Initially, \name{} undergoes a warm-up phase for \(T_0\) epochs. Subsequent epochs involve loss weight assignment and label update influenced by LID scores. To counteract error accumulation, \name{} integrates two augmented views for each sample, using their respective LID scores to guide weight calculation and label update.

\section{The Design of of Equation 3.13-3.15}
The design of \(w_{i,c}\), \(w_{i,h}\), \(w_{i,n}\) in equations aims to ensure that the sum of  \( w_{i,c} \), \( w_{i,h} \), and \( w_{i,n} \) equals 1.
\begin{proof}
    Without loss of generality, assume that \(w_{i,1} > w_{i,2}\). Then:
    \begin{align*}
        w_{i,c} &= w_{i,2}, \\
        w_{i,h} &= w_{i,1} - w_{i,2}, \\
        w_{i,n} &= 1 - w_{i,2}. \\
    \end{align*}
    Hence \(w_{i,c} + w_{i,h} + w_{i,n} = 1\).
\end{proof}

\section{The Design of Equation 3.36 and 3.37}

The design of \((2 - \Delta\hat{y}^k_i)/2\) and \((2 - \Delta\tilde{y}^k_i)/2\) terms in equations aims to map both \(\Delta\hat{y}^{k_i}\) and \(\Delta\tilde{y}^{k_i}\) into the interval \([0,1]\). This is based on the fact that the range of \(\Delta\) is \([0,2]\).

\begin{proof}
1. Consider vectors \(y = [y_1, y_2, ..., y_n]\) and \(u = [u_1, u_2, ...,u_n]\), where \( y_i \in [0,1], u_i \in [0,1]. \sum_i^n{y_i} = 1\), and \(\sum_i^n{u_i} = 1\). Define \(\Delta\) as \(\Delta = \sum_{i=1}^n|y_i - u_i|\).

2. Without loss of generality, assume that \(y_i \geq u_i\) for \(i \in [1, 2, ..., m]\) and \(y_j < u_j\) for \(j \in [m+1, m+2, ..., n]\). Then:
    \begin{align*}
        \Delta &= \sum_{i=1}^m{(y_i - u_i)} + \sum_{j=m+1}^n{(u_j - y_j)}.
    \end{align*}

3. Rearranging and utilizing the fact that the summation of each vector is 1, we deduce:
    \begin{align*}
        \Delta &= \sum_{i=1}^m{y_i} - \sum_{i=1}^{m}{u_i} + \sum_{j=m+1}^n{u_j} - \sum_{j=m+1}^{n}{y_j},
    \end{align*}
    \begin{align*}
  \sum_{i=1}^m{y_i} + \sum_{j=m+1}^{n}{y_j}  &= \sum_{i=1}^m{u_i} + \sum_{j=m+1}^{n}{u_j}, \\
        \sum_{i=1}^m{y_i} - \sum_{i=1}^{m}{u_i} &= \sum_{j=m+1}^n{u_j} - \sum_{j=m+1}^{n}{y_j}, \\
  \Delta &= 2(\sum_{i=1}^m{y_i} - \sum_{i=1}^{m}{u_i}).
    \end{align*}

4. Since \(\sum_{i=1}^m{y_i} \in [0,1],  \sum_{i=1}^m{u_i} \in [0,1]\) and \(\sum_{i=1}^m{y_i} \geq \sum_{i=1}^m{u_i} \geq 0\), it is implied that \((\sum_{i=1}^m{y_i} - \sum_{i=1}^m{u_i})\in [0,1]\).

5. Hence, \(\Delta \in [0, 2]\) and consequently, \(\frac{2 - \Delta}{2} \in [0,1]\).
\end{proof}

\section{Experiment Setup}
All experiments are executed using A100 GPUs and PyTorch 1.13.1. We use an AdamW \cite{loshchilov2017decoupled_adamw} optimizer with a learning rate of 0.001 and a weight decay of 0.001. The training epochs are 200 and the batch size is 128. \name{} first warms up for 15 epochs, during the warm-up stage, \name{} is optimized with cross entropy loss from two views, without weight assignment or label update. 

Inspired by \cite{li2023disc}, for \name{}, we employ two separate augmentation strategies to produce two views. The first approach involves random cropping combined with horizontal flipping, and the second incorporates random cropping, horizontal flipping, and RandAugment \cite{Tan_Xia_Wu_Li_2021_colearning}.  The value of \(\epsilon^{W}_{\mathrm{low}}\) and \(\epsilon^{U}_{\mathrm{low}}\) are both 0.001. The value of \(\epsilon^{W}_{\mathrm{high}}\) and \(\epsilon^{U}_{\mathrm{high}}\) start at values of 0.05 and 0.5 respectively, linearly increase to 1.0 in 30 epochs. The value of \(\epsilon_{k}\) is fixed at 0.1. The values of \(\lambda^{*}\) and \(\lambda_{cons}\) are 0.5 and 10 respectively.

In real-world applications, the noise ratio and pattern are often unknown. Earlier studies \cite{han2018co, yu2019does, li2019dividemix} assumed the availability of prior knowledge about the noise ratio or pattern, and they based their hyper-parameter settings on these assumptions. Recent works \cite{li2023disc} contend that such information is typically inaccessible in practice. Even though these works claim not to rely explicitly on noise information, their hyper-parameters still change as the noise ratio and type shift. For the sake of a fair comparison, we re-evaluated certain methods (indicated by a * symbol after the method name) using their open-sourced code. However, if these methods originally assumed unknown noise ratios or types, we kept their hyper-parameters consistent. The hyper-parameter settings we adopted were based on their medium noise ratio configurations for each noise type (50\% symmetric noise, 40\% asymmetric noise, 40\% instance dependent noise). We executed each method five times and recorded the average of the top three accuracy scores obtained during the training process. We employ a distinct backbone model for instance-dependent noise and real-world noise to ensure our results are comparable with those presented in \cite{liu2022robust, li2023disc}.